\documentclass{article}
\usepackage{arxiv}
\usepackage[utf8]{inputenc}
\usepackage[T1]{fontenc}
\usepackage{microtype}
\usepackage{amsmath}
\usepackage{amssymb}
\usepackage{booktabs}
\usepackage{enumitem}
\usepackage{graphicx}
\usepackage{flafter}
\usepackage[authoryear]{natbib}
\usepackage[hyphens]{url}
\usepackage{xcolor}
\usepackage{hyperref}
\hypersetup{
  hidelinks,
  pdftitle={Graph-Structured Rubrics: Compiling Rubrics into Typed Evaluation Graphs for LLM Judges},
  pdfauthor={Xi Chen, Jie Mu, Mo Xuan, Qun Shao}
}
\definecolor{gaincolor}{RGB}{0,110,70}
\newcommand{\gain}[1]{\rlap{\hspace{0.25em}\textcolor{gaincolor}{\scriptsize\bfseries +#1}}}

\title{Graph-Structured Rubrics: Compiling Rubrics into Typed Evaluation Graphs for LLM Judges}
\author{
  Xi Chen, Jie Mu, Mo Xuan, Qun Shao\\
  Ant Group\\
  \texttt{ct539484@antgroup.com}
}
\date{August 2026}

\begin{document}
\maketitle
\thispagestyle{firstpage}

\begin{abstract}
Rubric-based evaluators commonly treat rubrics as prompt context or flat
criteria: they specify what to judge but leave criterion composition implicit,
even when natural-language rules state it. We introduce
\emph{Graph-Structured Rubrics} (GSR), which compiles a rubric into a
response-independent typed evaluation graph before observing responses.
Criterion nodes elicit judgments; transformation, reduction, and gating
operators compose them through named ports; and a task-specific output mapping,
termed Readout, converts the unique sink into a score or preference.
Compilation rejects malformed or type-incompatible graphs. Pointwise evaluation
judges rubric dimensions separately before graph aggregation; pairwise
evaluation reuses the graph with one judgment for each candidate under every
criterion. Under GPT-OSS-120B, GSR improves exact score agreement by
0.62--6.75 percentage points over Prometheus-style scoring on four pointwise
datasets and achieves the numerically highest end-to-end pairwise accuracy on
two preference benchmarks under native tie and abstention policies.
\end{abstract}

\section{Introduction}

Rubrics have become a common interface for LLM evaluation across direct
scoring and pairwise comparison, yet many systems still treat them as prompt
context or flat collections of criteria. Such representations specify what to
judge but not how criterion-level judgments should interact to produce a score
or preference. LLM-based judges make such evaluation scalable and can achieve
substantial agreement with human judgments on open-ended tasks
\citep{zheng2023judging,kim2024prometheus2}.

\begin{figure}[!htbp]
\centering
\includegraphics[width=0.5\textwidth]{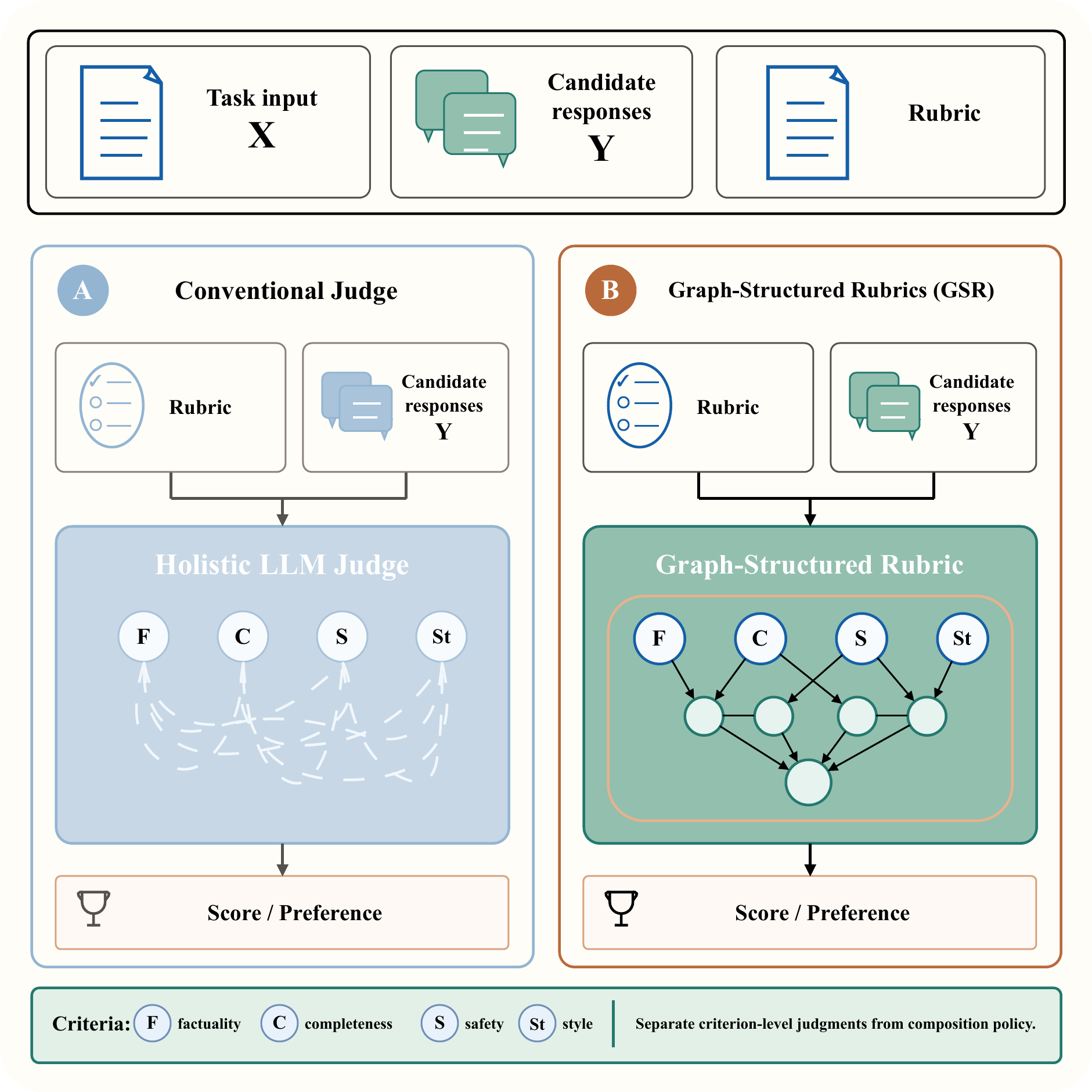}
\caption{Natural-language rules leave hierarchical composition implicit within
model inference; GSR routes criterion-level judgments through an explicit
graph before producing the final score or preference.}
\label{fig:gsr-introduction}
\end{figure}

As evaluated tasks become more heterogeneous, however, high aggregate
agreement with human ratings does not guarantee procedural stability. Prior
studies document sensitivity to candidate order, output length, evaluator
familiarity, anchoring, and even confusion between named quality criteria
\citep{wang2023fair,dubois2024length,stureborg2024inconsistent,hu2024confusing}.
Rubrics address the criterion-specification component of this problem by
making relevant requirements explicit. Accordingly, recent benchmarks and
evaluators have adopted human-authored criteria, generated rubrics, or
instance-specific checklists
\citep{zhou2026rubricbench,liu2025openrubrics,lee2024checkeval,tick2024}.
Because rubric-based judgments increasingly inform benchmark reporting,
reward-model data construction, and model-selection workflows, the way a
rubric is composed affects not only accuracy but also reproducibility and
auditability.

An important gap remains after criterion-level judgments have been produced:
their composition can still be implicit and unauditable. Writing rules in the
prompt does not remove this problem because the LLM must execute them
internally. Leaving these rules inside model inference creates opportunities
for applicability checks to be omitted, reductions or gates to be applied out
of order, or score caps to be enforced inconsistently. Moreover, rubric rules
are often hierarchical: criterion judgments feed intermediate reductions, whose
outputs feed gates, constraints, and final Readout. This hierarchy defines an
execution policy rather than additional criteria. To make that policy explicit,
GSR represents it as a directed acyclic graph and executes deterministic
operators in topological order, exposing the decision path shown in
Figure~\ref{fig:gsr-introduction}.

GSR therefore compiles each rubric before observing candidate responses. The
program contains criterion nodes, deterministic operators, named ports, and a
unique sink; compilation rejects cycles, missing ports, arity violations, and
type-incompatible routes. At evaluation time, language models interpret
semantic criteria, while the fixed graph controls routing, aggregation,
non-compensatory constraints, and Readout in topological order. The policy is
inspectable, and its execution is replayable from an audit trace that records
the decision path.

We evaluate the framework in pointwise and pairwise settings. The experiments
test whether compiled rubric graphs yield higher agreement than strong
baselines, whether the gains transfer across judge backbones, and whether they
arise from graph composition rather than direct scoring or flat aggregation.
Our contributions are:
\begin{itemize}[leftmargin=*]
    \item To our knowledge, we introduce the first response-independent
    compilation framework that turns a rubric specification into a typed
    cross-criterion evaluation DAG with
    rubric-derived criterion nodes, operator nodes, named ports, and a unique sink.
    \item We define execution semantics in which criterion-level judgments flow
    through deterministic \textsc{Transform}, \textsc{Reduce}, and
    \textsc{Gate} operators in topological order. Static validation rejects
    malformed graphs. Task-specific Readout produces pointwise or pairwise
    decisions, while audit traces enable deterministic replay of composition.
    \item We demonstrate the unified interface across four pointwise datasets
    and two pairwise preference benchmarks. Under GPT-OSS-120B, GSR achieves the
    numerically highest exact score agreement on all four pointwise datasets and
    the numerically highest end-to-end pairwise accuracy on both pairwise
    datasets; controlled ablations further show higher exact score agreement
    than direct scoring and weighted aggregation on every pointwise dataset.
\end{itemize}

\section{Related Work}

\paragraph{From holistic judges to explicit criteria.}
LLM judges produce direct scores or pairwise preferences, but their outputs are
sensitive to candidate position, response length, evaluator familiarity,
distribution shift, and anchoring
\citep{zheng2023judging,wang2023fair,dubois2024length,stureborg2024inconsistent}.
G-Eval adds structured reasoning \citep{liu2023geval}, JudgeLM uses swap and
reference augmentation \citep{zhu2023judgelm}, and Prometheus models support
custom criteria in direct and pairwise assessment
\citep{kim2024prometheus,kim2024prometheus2}. Debate and judge panels introduce
deliberation or evaluator diversity
\citep{chan2023chateval,verga2024juries}. Other work exposes finer-grained
criteria: BiGGen uses instance criteria \citep{kim2025biggen};
CheckEval, TICK, and RocketEval use checklist questions
\citep{lee2024checkeval,tick2024,wei2025rocketeval}; FLASK uses skill scores
\citep{ye2024flask}; and LMUnit treats criteria as unit tests
\citep{saadfalcon2025lmunit}. These approaches improve judgment production or
decomposition, but generally leave cross-criterion composition in prompts or
fixed aggregation rather than an explicit typed program.

\paragraph{Rubric construction, aggregation, and calibration.}
EvalLM supports application-specific criteria, while AutoCalibrate generates
and selects criteria against human labels
\citep{kim2023evallm,liu2024calibrating}. LLM-Rubric learns to aggregate
multidimensional rubric outputs into calibrated predictions, whereas Praetor
supports instance-level criteria in both pointwise and pairwise settings
\citep{hashemi2024llmrubric,leng2025praetor}. RubricBench, OpenRubrics, and
recursive rubric-refinement methods study rubric construction
\citep{zhou2026rubricbench,liu2025openrubrics,shen2026rethinking}, while
Autorubric unifies criterion design, judge ensembles, aggregation, and
calibration practices \citep{rao2026autorubric}. These works address criterion
selection, judgment elicitation, or score calibration. GSR assumes an available
rubric and gives executable semantics to its cross-criterion composition.

\paragraph{Graphs in evaluation and the remaining gap.}
Graph structures have been used in evaluation, but at different abstraction
boundaries. DAGMetric requires manual authoring of decision DAGs
\citep{confidentai2026dagmetric}; AgentEval derives graphs from agent execution
traces rather than rubric criteria \citep{guo2026agenteval}; and OpenRS
instantiates response-adaptive pairwise meta-rubrics and externally aggregates
criterion-wise preferences \citep{jia2026openrs}. RULERS instead compiles
criteria into locked executable specifications with deterministic evidence
verification and post-hoc score calibration \citep{hong2026rulers}. None of
these systems defines the specific abstraction studied here: a rubric compiled,
before observing candidate responses, into a response-independent typed
cross-criterion graph shared by pointwise and pairwise evaluation. GSR
addresses this gap by compiling operators and a task-specific Readout, statically
validating the resulting graph, and making its execution replayable.

\section{Graph-Structured Rubrics}
\label{sec:method}

GSR turns a hierarchical rubric policy into a graph program. Criterion nodes
produce semantic judgments, operator nodes encode reductions and constraints,
and edges fix their dependencies. Deterministic topological execution applies
the declared rules in order rather than asking the LLM to reconstruct that
order during every decision. For example, factuality and completeness can be
reduced before a safety flag gates the result, preventing later positive
evidence from overriding the cap. Pointwise scoring and pairwise preference
share these node and operator semantics and differ only in task-specific
Readout.

\begin{figure}[!htbp]
\centering
\includegraphics[width=\textwidth]{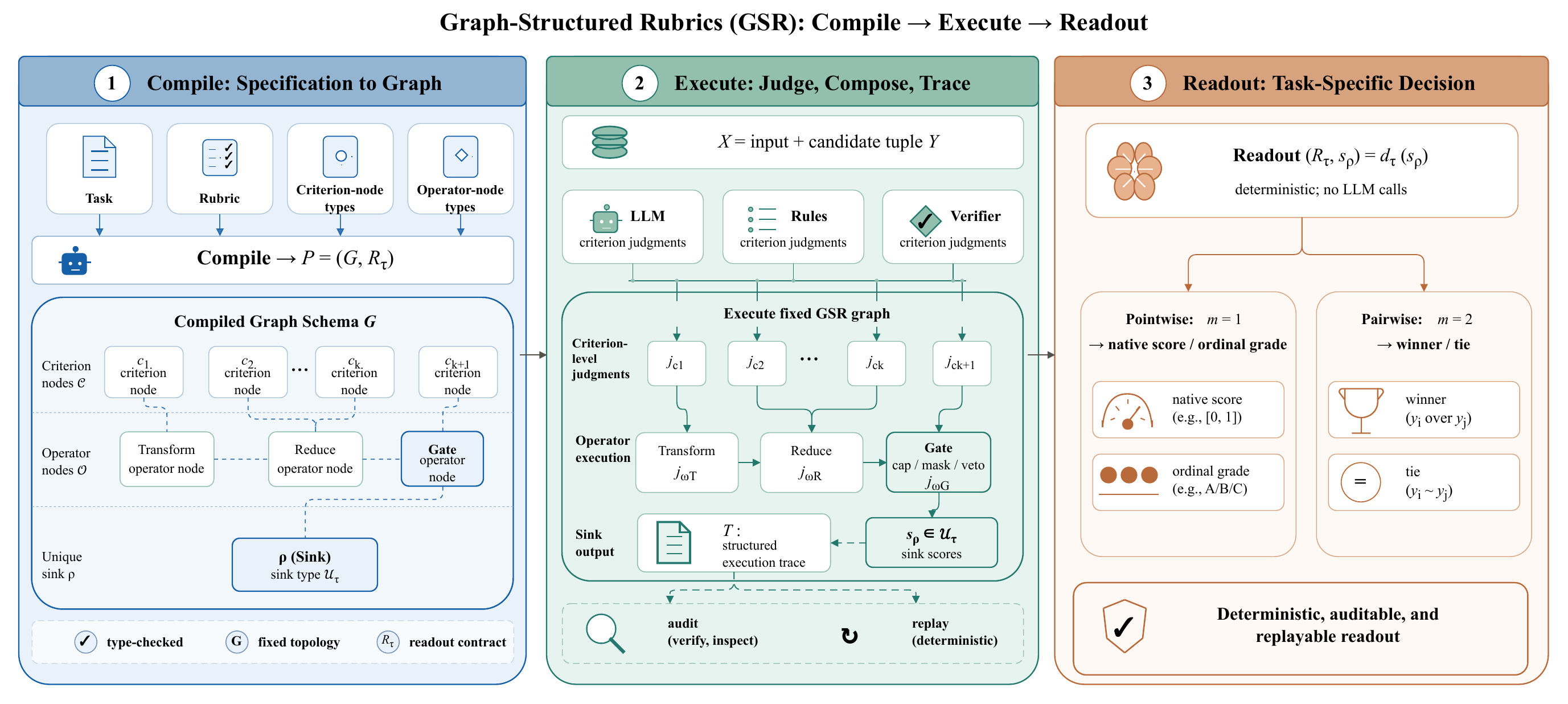}
\caption{GSR procedure. Compile fixes the graph before observing candidate responses;
Execute runs criterion and operator nodes; Readout maps sink scores to the
task output.}
\label{fig:gsr-overview}
\end{figure}

\paragraph{Problem setup.}
GSR fixes an instance-specific composition policy before observing the responses
it will judge. The compile-time specification
$S=(x,r,z,\tau)$ contains the task input $x$, rubric $r$, optional
fixed reference material $z$, and task contract $\tau$, but excludes candidate responses
and gold labels. The contract fixes the number of candidates $m$, a closed
internal quality-score interval $\mathcal U_\tau\subset\mathbb R$, output space
$\mathcal A_\tau$, and quantization and tie policies. The execution stage
subsequently receives $X=(x,Y)$, where
$Y=(y_1,\ldots,y_m)$ is an ordered candidate tuple with stable
identifiers. This partition is enforced at the invocation boundary:
compilation cannot access $Y$, while execution cannot alter the accepted
program.

\paragraph{Evaluation procedure.}
GSR has a three-stage interface. Given a graph language $\Lambda$,
Compile turns $S$ into a response-independent program,
\begin{equation}
P=\mathrm{Compile}(S;\Lambda)=(G,R_\tau),
\label{eq:overview-compile}
\end{equation}
where $G$ is the rubric graph and $R_\tau$ is a readout contract.
Execute applies the program to $X$ and returns the score vector produced at the
unique sink and an audit trace $T$,
\begin{equation}
\begin{aligned}
(\mathbf s_\rho,T)&=\mathrm{Execute}(P,X),\\
\mathbf s_\rho&=\bigl((\operatorname{id}(y_i),s_{\rho,i})\bigr)_{i=1}^{m},\\
s_{\rho,i}&\in\mathcal U_\tau.
\end{aligned}
\label{eq:overview-execute}
\end{equation}
Readout then returns
\begin{equation}
o=\mathrm{Readout}(R_\tau,\mathbf s_\rho)
=d_\tau(\mathbf s_\rho).
\label{eq:overview-readout}
\end{equation}

\subsection{Compile: From a Rubric to a Graph Program}
\label{sec:compile-validate}

Compile receives the response-independent specification $S$ and a graph
language
\begin{equation}
\Lambda=(\mathbb T,\mathbb F,\mathbb H,\Pi),
\label{eq:graph-catalog}
\end{equation}
where $\mathbb T$ gives the value types, $\mathbb F$ gives the available graph
operators, $\mathbb H$ gives the available criterion-level judgment procedures, and
$\Pi$ gives task defaults and composition policies. This graph language fixes
what kinds of criterion-level judgments can be produced and how they can
be composed. The compiler therefore receives the rubric and the available graph
language, but not the candidate responses or gold labels.

\paragraph{LLM-guided graph synthesis and repair.}
Compile uses an LLM to synthesize a declarative program from $S$ under
$\Lambda$. The model is restricted to catalogued types, criterion procedures,
and operators, and emits nodes, named-port edges, parameters, and
$R_\tau$. A deterministic validator $V$ checks JSON parsing, acyclicity, sink
reachability, ports, arity, routing type compatibility, and readout compatibility. Writing
$\delta^{(t)}$ for its structured diagnostics, compilation follows
\begin{equation}
\begin{aligned}
P^{(0)}&=C_\phi(S,\Lambda),\\
(b^{(t)},\delta^{(t)})&=V(P^{(t)}),\\
P^{(t+1)}&=C_\phi(S,\Lambda,P^{(t)},\delta^{(t)})
\quad\text{if }b^{(t)}=0.
\end{aligned}
\label{eq:compile-repair}
\end{equation}
The first structurally valid program with $b^{(t)}=1$ is accepted; no manual
semantic screening or selection among valid candidates is performed. If none
is valid after the fixed repair budget $K_{\mathrm{rep}}$, the instance is recorded as a
compilation failure. Validation checks that the graph is executable and
type-consistent; whether the accepted graph captures the intended meaning of
the natural-language rubric is tested through downstream agreement. Compiler
and judge can use the same underlying LLM, as in our experiments, but they are
separate invocations with different inputs: the compiler receives no candidate
response, whereas the judge receives the frozen program and candidates during
Execute.

The output of Compile is a program $P=(G,R_\tau)$. The graph
\begin{equation}
\begin{aligned}
G&=(\mathcal{V},\mathcal{E},\rho),\\
\mathcal{V}&=\mathcal{C}\mathbin{\dot\cup}\mathcal{O},\qquad \rho\in\mathcal V,\\
\mathcal{E}&\subseteq
\mathcal{V}\times\mathcal{P}\times\mathbb N_{+}\times\mathcal{O}.
\end{aligned}
\label{eq:graph}
\end{equation}
contains criterion nodes $\mathcal C$, operator nodes $\mathcal O$,
type-checked routing edges $\mathcal E$, port names $\mathcal P$, and a unique
sink $\rho$. An edge
$(u,p,\ell,\omega)$ routes the judgment produced by predecessor node $u$ to
slot $\ell$ of named port $p$ of operator node $\omega$. The graph permits
criterion-to-operator and operator-to-operator edges; criterion nodes have no
incoming judgment-flow edges. We write
$v\leadsto\rho$ when node $v\in\mathcal V$ has a directed path to the sink
$\rho$, and require this condition for every node so that no criterion-level judgment is
unreachable. Each node declares an output type, each operator port declares an
input type, and compilation rejects cycles, missing ports, arity mismatches,
and type-incompatible routes.

Criterion-node semantics are task-invariant. Compile instantiates one node for
each semantic criterion selected from the rubric, and every criterion node
returns candidate-aligned judgments for all $m$ candidates. Pointwise
evaluation sets $m=1$, whereas pairwise evaluation sets $m=2$. Here $m$
counts candidates, not criteria: either regime may contain multiple criterion
nodes corresponding to different rubric dimensions. In both regimes, criterion
nodes represent rubric dimensions rather than native task labels; their
differences are confined to candidate arity, the task contract, and Readout.

Within $\mathbb F$, operators are grouped by the kind of composition they
perform. Let $\mathcal J$ denote judgment spaces and $\mathcal B$ Boolean flags:
a \textsc{Transform} operator maps $\mathcal J_a$ to $\mathcal J_b$, a
\textsc{Reduce} operator maps $\mathcal J_1\times\cdots\times\mathcal J_k$ to
$\mathcal J$, and a \textsc{Gate} operator maps $\mathcal J\times\mathcal B$ to
$\mathcal J$ by applying a declared cap, mask, or veto. \textsc{Gate} is
therefore an operator node in $\mathcal O$, not an additional node category
beyond $\mathcal C$ and $\mathcal O$.

Compile also attaches a readout contract
\begin{equation}
\begin{aligned}
R_\tau&=(\operatorname{Align}_m(\mathcal U_\tau),
\mathcal A_\tau,d_\tau),\\
d_\tau&:\operatorname{Align}_m(\mathcal U_\tau)\rightarrow\mathcal A_\tau.
\end{aligned}
\label{eq:readout-contract}
\end{equation}
Here $\operatorname{Align}_m(\mathcal U_\tau)$ denotes the arity-$m$ schema for
candidate-aligned quality scores; for an evaluated instance it is instantiated as
$((\operatorname{id}(y_i),s_{\rho,i}))_{i=1}^{m}$ with
$s_{\rho,i}\in\mathcal U_\tau$. The map $d_\tau$ alone performs native-scale
conversion, output quantization, and tie or forced-choice behavior. For
pairwise tasks, the annotated label belongs to $\mathcal A_\tau$
(for example, a winner or tie); the per-candidate scores are internal sink
values produced by the graph. The sink type must be an aligned sequence of $m$
quality scores in $\mathcal U_\tau$. The experiments instantiate this shared
contract for pointwise scoring and pairwise preference.

\subsection{Execute: Evaluate and Compose Judgments}
\label{sec:execute}

Execute applies the fixed graph to $X$ without changing its nodes, edges, or
parameters. Each criterion node $c\in\mathcal C$ produces a candidate-aligned
judgment vector
\begin{equation}
\mathbf j_c=\bigl((\operatorname{id}(y_i),j_{c,i})\bigr)_{i=1}^{m},
\qquad j_{c,i}\in\mathsf{Judg}(c).
\label{eq:criterion-execution}
\end{equation}
Here $\mathsf{Judg}(c)$ is the judgment space declared for criterion $c$. The
candidate-aligned vector can be produced by evaluating candidates separately or
jointly, but it must contain exactly one criterion-level judgment for each
candidate identifier. A criterion node never receives another node's judgment.
Criterion procedures include LLM calls, rule checks, verifiers, or human
annotations, but their outputs must conform to the criterion node's declared
judgment type.

After the criterion-level judgments are available, operators run in a topological order
of $G$. Each operator declares named input ports. For a port $p$ of operator
$\omega$, let $\operatorname{Pred}(\omega,p)=(v_1,\ldots,v_k)$ denote the
predecessor nodes whose edges target $p$, ordered by their edge slots. The input
received at that port is
\[
\mathbf j^{\mathrm{in}}_{\omega,p}
=(\mathbf j_{v_1},\ldots,\mathbf j_{v_k}).
\]
The operator then produces
\begin{equation}
\mathbf j_\omega=
f_\omega\!\left(
\bigl(p\mapsto\mathbf j^{\mathrm{in}}_{\omega,p}\bigr)_{p\in
\operatorname{ports}(\omega)};
\theta_\omega
\right).
\label{eq:operator-execution}
\end{equation}
The deterministic function $f_\omega$ receives a named, slot-ordered sequence
at each port and applies the fixed parameters $\theta_\omega$; positional
parameters are validated against the corresponding sequence length.
For instance, a \textsc{Gate:Cap} reads a quality judgment through
\texttt{base} and a Boolean failure judgment through \texttt{trigger}; the edge
itself carries no cap semantics.

Let $\mathbf j_v$ denote the candidate-aligned output vector of node $v$.
Operators preserve candidate identifiers, and the terminal node is type-checked
to return aligned quality scores. Thus all graph-level reductions and gates
have been applied when the sink scores are produced:
\begin{equation}
\mathbf s_\rho=\mathbf j_\rho=
\bigl((\operatorname{id}(y_i),s_{\rho,i})\bigr)_{i=1}^{m},
\quad s_{\rho,i}\in\mathcal U_\tau.
\label{eq:sink-score-vector}
\end{equation}
Each $s_{\rho,i}$ is the final internal quality score for $y_i$. Under the
fixed topological order, the audit trace records node identifiers, slot-ordered
inputs, outputs, operator parameters, and evidence references so that the
composition stage can be replayed from recorded criterion-level judgments.

\subsection{Readout: Decide from Candidate Scores}
\label{sec:readout}

Readout is fixed during Compile and makes no additional LLM call. It receives
$\mathbf s_\rho=((\operatorname{id}(y_i),s_{\rho,i}))_{i=1}^{m}$, checks
identifier coverage, finiteness, and domain membership, and applies the
deterministic task policy in $R_\tau$:
\begin{equation}
o=d_\tau(\mathbf s_\rho).
\label{eq:readout-decision}
\end{equation}
For pointwise evaluation ($m=1$), $d_\tau$ returns the native score or ordinal
label obtained by the deterministic conversion and quantization policies in
$\tau$. For
pairwise evaluation ($m=2$),
\begin{equation}
d_\tau(\mathbf s_\rho)=
\begin{cases}
 \operatorname{id}(y_1), & s_{\rho,1}>s_{\rho,2}+\epsilon_\tau,\\
 \operatorname{id}(y_2), & s_{\rho,2}>s_{\rho,1}+\epsilon_\tau,\\
 \operatorname{resolve}_\tau(\operatorname{id}(y_1),\operatorname{id}(y_2)),
 & |s_{\rho,1}-s_{\rho,2}|\leq\epsilon_\tau,
\end{cases}
\label{eq:pairwise-readout}
\end{equation}
where $\epsilon_\tau\geq0$ and $\operatorname{resolve}_\tau$ are fixed by the
task contract. The latter returns a tie, an abstention, or a candidate selected
by a declared deterministic forced-choice rule. Scores are comparable only
within the same program and task contract; GSR does not assume calibration
across rubrics or judge models.

Returning to the factuality--completeness--safety example, the graph computes
one quality score for each candidate after applying the safety cap, if any. In
a pointwise task, the same criterion graph emits one aligned sink score and
Readout quantizes it onto the native rubric scale. In a pairwise task, it emits
two aligned sink scores and Readout compares them under the task's tie policy.
Thus pointwise scoring and pairwise preference share criterion-node and
operator semantics; they differ only in candidate arity, task contract, and
Readout. Criterion-level judgments remain model-dependent, but their routing,
composition, gates, and final mapping are statically checked and replayable.

\section{Experiments}
\label{sec:experiments}

\subsection{Experimental Setup}

\paragraph{Evaluation questions.}
We evaluate ordinal pointwise scoring and A/B preference through the same
criterion graph. Pointwise judging produces one judgment per rubric dimension
and a native 1--5 score; pairwise judging produces two candidate-aligned
judgments per criterion before preference Readout. Only candidate arity and
task output change.

\paragraph{Datasets and baselines.}
Pointwise datasets are UltraFeedback--TruthfulQA (3,244 responses)
\citep{cui2024ultrafeedback,lin2022truthfulqa}, HelpSteer2 validation (1,038)
\citep{wang2024helpsteer2}, SummEval Relevance (1,600)
\citep{fabbri2021summeval}, and BiGGen (2,776) \citep{kim2025biggen}; their
heterogeneous 1--5 targets are termed \emph{reference scores}. Pairwise
datasets are MT-Bench (2,575 labeled pairs) \citep{zheng2023judging} and
RubricBench (1,147) \citep{zhou2026rubricbench}. Unless noted, GPT-OSS-120B is
the judge. Pointwise baselines are controlled, same-backbone adaptations of
Prometheus \citep{kim2024prometheus2}, G-Eval \citep{liu2023geval}, and FLASK
\citep{ye2024flask}; pairwise baselines add OpenRubric
\citep{liu2025openrubrics}, TICK \citep{tick2024}, and CheckEval
\citep{lee2024checkeval}.
Evaluation criteria are taken from public benchmark annotations or task
definitions. In pointwise experiments, each benchmark is judged with its
corresponding task criterion or released scoring rubric; in pairwise
experiments, MT-Bench uses a shared preference rubric and RubricBench uses
released instance-level checklist rubrics. For each benchmark, GSR and the
baselines receive the same available task information, including the input,
candidate response(s), and criterion/rubric text, with fixed reference material
when provided. Gold scores and preference labels are reserved for metric
computation, and baselines retain their native prompting and output formats.

\begin{table}[!htbp]
\centering
\small
\resizebox{\textwidth}{!}{%
\begin{tabular}{@{}llrrrrr@{}}
\toprule
\textbf{Dataset} & \textbf{Method}
& \textbf{Exact Agreement} $\uparrow$
& \textbf{Within-1 Accuracy} $\uparrow$
& \textbf{MAE} $\downarrow$ & \textbf{Pearson} $\uparrow$ & \textbf{Spearman} $\uparrow$ \\
\midrule
UF--TruthfulQA & Prometheus-style & 43.30 & 73.50 & 0.928 & 0.637 & 0.623 \\
 & G-Eval-style & 44.61 & 74.98 & 0.881 & 0.593 & 0.565 \\
 & FLASK-style & 39.02 & 68.39 & 1.062 & 0.661 & \textbf{0.669} \\
 & \textbf{GSR (Ours)} & \textbf{50.05}\gain{5.44} & \textbf{78.75}\gain{3.77} & \textbf{0.779}\gain{0.102} & \textbf{0.678}\gain{0.017} & 0.659 \\
\addlinespace[4pt]
HelpSteer2 & Prometheus-style & 38.97 & 77.38 & 0.928 & \textbf{0.514} & \textbf{0.444} \\
 & G-Eval-style & 35.16 & 77.14 & 0.976 & 0.504 & 0.438 \\
 & FLASK-style & 35.81 & 75.48 & 0.978 & 0.468 & 0.405 \\
 & \textbf{GSR (Ours)} & \textbf{43.75}\gain{4.78} & \textbf{80.81}\gain{3.43} & \textbf{0.845}\gain{0.083} & \textbf{0.514} & 0.427 \\
\addlinespace[4pt]
SummEval Relevance & Prometheus-style & 32.70 & 83.73 & 0.846 & \textbf{0.441} & \textbf{0.424} \\
 & G-Eval-style & 28.28 & 79.04 & 0.942 & 0.429 & 0.405 \\
 & FLASK-style & 28.83 & 72.55 & 1.068 & 0.301 & 0.286 \\
 & \textbf{GSR (Ours)} & \textbf{33.32}\gain{0.62} & \textbf{86.75}\gain{3.02} & \textbf{0.804}\gain{0.042} & 0.392 & 0.386 \\
\addlinespace[4pt]
BiGGen & Prometheus-style & 43.52 & 79.73 & 0.854 & 0.611 & 0.588 \\
 & G-Eval-style & 40.83 & 79.12 & 0.890 & 0.589 & 0.565 \\
 & FLASK-style & 38.27 & 79.93 & 0.891 & 0.570 & 0.546 \\
 & \textbf{GSR (Ours)} & \textbf{44.51}\gain{0.99} & \textbf{80.49}\gain{0.56} & \textbf{0.833}\gain{0.021} & \textbf{0.618}\gain{0.007} & \textbf{0.591}\gain{0.003} \\
\bottomrule
\end{tabular}
}
\caption{Main pointwise comparison under GPT-OSS-120B, averaged over six runs
(green: improvement vs.\ the best baseline for each metric).}
\label{tab:main-comparison}
\end{table}

\begin{figure}[!htbp]
\centering
\includegraphics[width=0.485\textwidth]{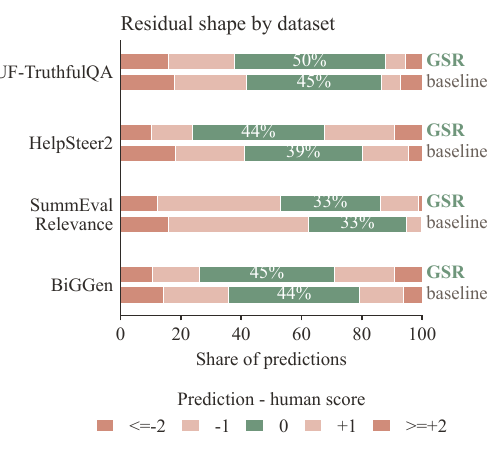}
\hfill
\includegraphics[width=0.485\textwidth]{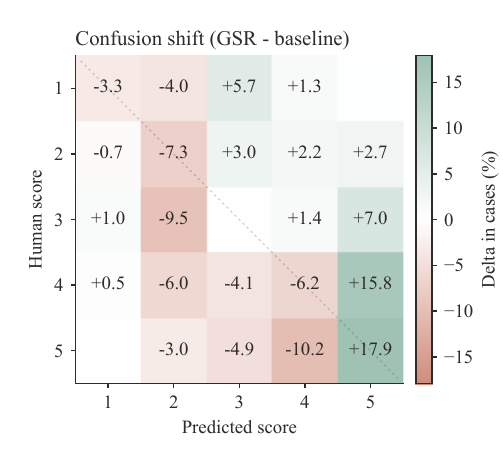}
\caption{Six-run pointwise error geometry versus the strongest Exact Agreement baseline:
pooled residuals (left) and mean row-normalized confusion shift (right).
The baseline is G-Eval-style for UF--TruthfulQA and Prometheus-style otherwise.}
\label{fig:pointwise-diagnostics}
\end{figure}

\begin{table}[!htbp]
\centering
\small
\begin{tabular*}{\textwidth}{@{\extracolsep{\fill}}llrrrrr@{}}
\toprule
\textbf{Dataset} & \textbf{Method}
& \textbf{Pairwise Accuracy} $\uparrow$
& \textbf{Valid Accuracy} $\uparrow$
& \textbf{Coverage} $\uparrow$
& \textbf{Invalid Rate} $\downarrow$
& \textbf{Tie Rate} $\downarrow$ \\
\midrule
MT-Bench & OpenRubric & 79.86 & 79.86 & 100.00 & 0.00 & 0.00 \\
 & TICK & 54.10 & 84.36 & 64.13 & 1.21 & 34.66 \\
 & CheckEval & 65.28 & 85.05 & 76.74 & 0.52 & 22.74 \\
 & \textbf{GSR (Ours)} & \textbf{80.63}\gain{0.77} & 80.73 & 99.87 & 0.13 & 0.00 \\
\addlinespace[4pt]
RubricBench & OpenRubric & 83.35 & 84.30 & 98.87 & 1.13 & 0.00 \\
 & TICK & 68.98 & 88.38 & 78.04 & 2.18 & 19.78 \\
 & CheckEval & 70.49 & 88.59 & 79.57 & 0.86 & 19.57 \\
 & \textbf{GSR (Ours)} & \textbf{83.62}\gain{0.28} & 83.70 & 99.91 & 0.09 & 0.00 \\
\bottomrule
\end{tabular*}
\caption{Pairwise comparison under GPT-OSS-120B, averaged over six runs
(green: $\Delta$ Accuracy vs.\ OpenRubric).}
\label{tab:pairwise-comparison}
\end{table}

\begin{figure}[!b]
\centering
\includegraphics[width=0.5\textwidth]{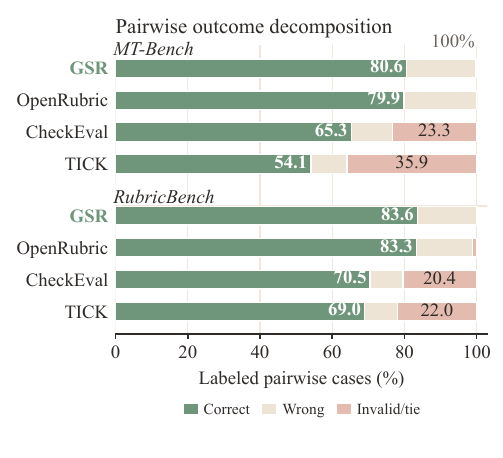}
\caption{Pairwise outcomes under native decision policies. Invalid outputs
and ties produced by baselines are grouped as non-decisions. Valid-only accuracy can
remain high despite lower end-to-end accuracy.}
\label{fig:pairwise-outcomes}
\end{figure}

\paragraph{Implementation protocol.}
Each instance is recompiled in a separate API call; the first
validator-approved program within $K_{\mathrm{rep}}$ repairs is accepted
without human selection and frozen for execution. Run-to-run variation
therefore includes compilation and judging nondeterminism.

\paragraph{Metrics and provenance.}
Pointwise Exact Agreement, Within-1 Accuracy, MAE, Pearson, and Spearman are
computed from predicted scores. Pairwise Coverage counts A/B decisions, Valid
Accuracy conditions on them, and Pairwise Accuracy counts ties and invalid
outputs as errors. Invalid Rate includes compile/parse failures and
abstentions. GSR treats unresolved comparisons as abstentions; ties produced by
baselines remain ties. Unless otherwise noted, reported tables use arithmetic
means over six complete runs. Predictions retain dataset, model, protocol, and
reference provenance, and no tie or abstention is forced post hoc into a choice.
When standard deviations are summarized in text, they are sample standard
deviations over the same six run-level primary metrics.

\subsection{Main Results}

Tables~\ref{tab:main-comparison} and~\ref{tab:pairwise-comparison} report
pointwise and pairwise results under GPT-OSS-120B. Exact Agreement and
end-to-end Pairwise Accuracy are the respective primary metrics. These metrics
test the final decision produced after criterion-level judgments have been
composed, and therefore evaluate the complete GSR pipeline rather than the
accuracy of any individual criterion node.

For pointwise scoring, GSR gives the best Exact Agreement, Within-1 Accuracy,
and MAE in every dataset block. Relative to the strongest baseline by Exact
Agreement, its gains are 5.44 points on UF--TruthfulQA, 4.78 on HelpSteer2,
0.62 on SummEval Relevance, and 0.99 on BiGGen. The corresponding Within-1
gains are 3.77, 3.43, 3.02, and 0.56 points, while MAE falls by 0.102, 0.083,
0.042, and 0.021. The smaller Exact margins on SummEval Relevance and BiGGen
indicate that GSR is competitive rather than decisively separated there. Across
the four datasets, the common result is therefore not a uniformly large margin,
but a consistent advantage on the final discrete score produced by the graph
and Readout. This is compatible with GSR's intended role of making
cross-criterion dependencies explicit after criterion judgments are obtained.
Because the main comparison changes the full evaluation procedure, the
same-trace ablation below provides the more direct test of graph composition.
Across the four GSR rows, the six-run standard deviations for Exact Agreement
are 0.45--0.84 points.

Figure~\ref{fig:pointwise-diagnostics} localizes these changes. GSR increases
zero residuals most clearly on UF--TruthfulQA and HelpSteer2, but the confusion
shift is not uniformly favorable. For reference score 5, mass moves from
mispredictions of 4 to the correct score 5; for reference score 4, some mass instead
moves from 4 to 5. This mixed high-score redistribution is consistent with
Exact Agreement and MAE improving more uniformly than Pearson or Spearman.
Thus, explicit graph execution should not be read as a generic correction that
moves every prediction toward its reference. It changes how criterion-level
outputs cross the rubric's final decision boundaries, and those changes can
help one reference level while hurting a neighboring level. GSR therefore
provides a controlled composition path, not a uniform correction of the score
distribution.

The same criterion-node interface transfers to pairwise decisions without
architectural modification. GSR attains the highest end-to-end Pairwise Accuracy
at 99.87--99.91\% coverage; TICK and CheckEval have higher valid-only accuracy
but 19.57--34.66\% tie rates, which Figure~\ref{fig:pairwise-outcomes} counts
as non-decisions. The GSR Pairwise Accuracy standard deviation is 0.30 points on
MT-Bench and 0.51 points on RubricBench.

\subsection{Ablation Study}

Direct predicts one holistic score. Weighted aggregation applies a flat
weighted rule to the same criterion-level judgments as GSR, whereas full GSR
applies compiled operators and task-level quantization. This isolates the
effect of graph composition.

\begin{table}[!htbp]
\centering
\small
\setlength{\tabcolsep}{1.6pt}
\begin{tabular*}{\linewidth}{@{\extracolsep{\fill}}lccccc@{}}
\toprule
\textbf{Variant} & \textbf{Exact Agreement} $\uparrow$ & \textbf{Within-1 Accuracy} $\uparrow$
& \textbf{MAE} $\downarrow$ & \textbf{Pearson} $\uparrow$ & \textbf{Spearman} $\uparrow$ \\
\midrule
\multicolumn{6}{@{}l}{\textit{BiGGen}} \\
Direct & 41.00 & 79.13 & 0.889 & 0.589 & 0.562 \\
Weighted aggregation & 41.91 & \textbf{80.93} & 0.836 & 0.615 & 0.590 \\
\textbf{GSR composition} & \textbf{44.51}\gain{2.60} & 80.49 & \textbf{0.833}\gain{0.003} & \textbf{0.618}\gain{0.003} & \textbf{0.591}\gain{0.001} \\
\addlinespace[5pt]
\multicolumn{6}{@{}l}{\textit{UF--TruthfulQA}} \\
Direct & 43.04 & 75.49 & 0.897 & 0.612 & 0.586 \\
Weighted aggregation & 44.26 & 77.03 & 0.851 & 0.664 & 0.649 \\
\textbf{GSR composition} & \textbf{50.05}\gain{5.79} & \textbf{78.75}\gain{1.72} & \textbf{0.779}\gain{0.072} & \textbf{0.678}\gain{0.014} & \textbf{0.659}\gain{0.010} \\
\addlinespace[5pt]
\multicolumn{6}{@{}l}{\textit{HelpSteer2}} \\
Direct & 35.79 & 77.16 & 0.963 & \textbf{0.522} & \textbf{0.451} \\
Weighted aggregation & 40.49 & \textbf{82.43} & \textbf{0.844} & 0.521 & 0.446 \\
\textbf{GSR composition} & \textbf{43.75}\gain{3.26} & 80.81 & 0.845 & 0.514 & 0.427 \\
\addlinespace[5pt]
\multicolumn{6}{@{}l}{\textit{SummEval Relevance}} \\
Direct & 24.07 & 75.92 & 1.022 & \textbf{0.454} & \textbf{0.442} \\
Weighted aggregation & 32.96 & 85.74 & 0.817 & 0.401 & 0.382 \\
\textbf{GSR composition} & \textbf{33.32}\gain{0.36} & \textbf{86.75}\gain{1.01} & \textbf{0.804}\gain{0.013} & 0.392 & 0.386\gain{0.004} \\
\bottomrule
\end{tabular*}
\caption{Six-run scoring and composition ablation. Full GSR rows reuse the
predictions in Table~\ref{tab:main-comparison}; weighted-aggregation
predictions are recomputed from the same criterion-level judgment traces
(green: improvement vs. weighted aggregation; MAE annotations show reductions).}
\label{tab:readout-ablation}
\end{table}

GSR improves Exact Agreement over Direct by 3.51--9.25 points. Against weighted
aggregation, the gains are 2.60 points on BiGGen, 5.79 on UF--TruthfulQA, 3.26
on HelpSteer2, and 0.36 on SummEval Relevance. Because the latter comparison
reuses the same criterion-level traces, these differences isolate composition
and Readout rather than criterion elicitation: weighted aggregation collapses
criterion-wise evidence in one flat rule, whereas GSR routes the same evidence
through compiled \textsc{Transform}, \textsc{Reduce}, and \textsc{Gate}
operators before task-level Readout.

The GSR-composition rows reuse the same six runs as the GSR rows in
Table~\ref{tab:main-comparison}; their Exact Agreement standard deviations are
therefore 0.45--0.84 points across datasets. The gain varies from 0.36 to 5.79
points and does not identify which operator or dependency accounts for the
variation, but it supports the limited conclusion that the compiled graph can
change final ordinal boundaries while criterion-level inputs are held fixed.
The flat weighted variant still has better Within-1 Accuracy on BiGGen and
HelpSteer2, a marginally lower MAE on HelpSteer2, and sometimes stronger
correlation, so GSR improves exact ordinal selection here but does not dominate
a flat rule on all secondary metrics.

\subsection{Cross-Model Sensitivity}

\begin{table}[!htbp]
\centering
\small
\setlength{\tabcolsep}{1.8pt}
\begin{tabular*}{\linewidth}{@{\extracolsep{\fill}}lccccc@{}}
\toprule
\textbf{Method} & \textbf{Exact Agreement} $\uparrow$ & \textbf{Within-1 Accuracy} $\uparrow$
& \textbf{MAE} $\downarrow$ & \textbf{Pearson} $\uparrow$ & \textbf{Spearman} $\uparrow$ \\
\midrule
\multicolumn{6}{@{}l}{\textit{Qwen3.5-35B-A3B}} \\
Prometheus-style & 41.51 & 77.44 & 0.913 & \textbf{0.503} & 0.426 \\
G-Eval-style & 36.94 & 73.68 & 1.016 & 0.486 & 0.424 \\
FLASK-style & 39.03 & 77.22 & 0.932 & 0.461 & 0.393 \\
\textbf{GSR (Ours)} & \textbf{41.76}\gain{0.25} & \textbf{78.16}\gain{0.72} & \textbf{0.878}\gain{0.035} & 0.500 & \textbf{0.427}\gain{0.001} \\
\addlinespace[5pt]
\multicolumn{6}{@{}l}{\textit{GLM-4.7}} \\
Prometheus-style & \textbf{37.14} & 71.06 & 1.075 & \textbf{0.489} & \textbf{0.436} \\
G-Eval-style & 36.32 & 71.73 & 1.057 & 0.476 & 0.417 \\
FLASK-style & 34.52 & 72.17 & 1.056 & 0.452 & 0.399 \\
\textbf{GSR (Ours)} & 36.10 & \textbf{74.64}\gain{3.58} & \textbf{1.004}\gain{0.071} & \textbf{0.489} & 0.425 \\
\bottomrule
\end{tabular*}
\caption{Six-run cross-model sensitivity on HelpSteer2 (green: improvement
vs. Prometheus-style; MAE annotations show reductions).}
\label{tab:cross-model-sensitivity}
\end{table}

On HelpSteer2 with Qwen3.5-35B-A3B, GSR exceeds Prometheus-style by 0.25 points
in Exact Agreement and 0.72 points in Within-1 Accuracy, while reducing MAE by
0.035. With GLM-4.7, its Exact Agreement is 1.04 points lower, but its Within-1
Accuracy is 3.58 points higher and its MAE is lower by 0.071. The graph
interface therefore transfers structurally, but its metric profile is not
backbone-invariant. GSR fixes dependencies, operator order, and Readout; it
does not replace or correct the semantic judgments produced at criterion
nodes. The Qwen result preserves a small Exact advantage, whereas the GLM
comparison reverses despite better Within-1 Accuracy and MAE\@. Explicit
composition controls how judgments are combined, but remains conditioned on
the judgments entering the graph. The corresponding GSR Exact Agreement
standard deviations are 0.74 points for Qwen3.5-35B-A3B and 0.87 points for
GLM-4.7.

\section{Conclusion}

GSR composes criterion-level judgments through a typed graph shared by
pointwise and pairwise evaluation. Under GPT-OSS-120B, GSR is
numerically strongest on the primary end-to-end metric in all six evaluated
datasets, while ablations support the value of graph composition and
cross-model results bound that value by backbone dependence. GSR offers a
programmable alternative whose composition step is trace-replayable.
Generative AI assisted with language, formatting, and consistency checks; the
authors verified and remain responsible for all claims, results, and citations.

\begingroup
\small
\setlength{\bibsep}{2pt plus 0.5pt}
\bibliographystyle{plainnat}
\bibliography{references}
\endgroup

\end{document}